\documentclass[preprint,1p,11pt]{elsarticle}

\usepackage{amssymb}
\usepackage{subfigure}
\usepackage{lineno}
\usepackage{color}
\usepackage{ulem}
\usepackage{lineno}

\journal{Pattern Recognition Letters}

\begin{document}

\begin{frontmatter}

\title{Exponentially Weighted Moving Average Charts for Detecting Concept Drift}

\author[ic]{Gordon J.~Ross\corref{cor1}}
\ead{gordon.ross03@imperial.ac.uk}
 
\author[ic]{Niall M.~Adams}
\ead{n.adams@ic.ac.uk}
 
\author[ic]{Dimitris K.~Tasoulis}
\ead{d.tasoulis@ic.ac.uk}

\author[ic]{David J. ~Hand}
\ead{d.hand@ic.ac.uk}

\cortext[cor1]{Author for correspondence, Email: gordon.ross03@ic.ac.uk, Tel:
+44(0)2075940990, Fax: +44(0)2075940923}

\address[ic]{Department of Mathematics, Imperial College, London SW7
2AZ, UK}

%\textcolor{red}{text}

\begin{abstract}.Classifying streaming data
requires the development of methods
which are computationally efficient and able to cope with changes in the
underlying distribution of the stream, a phenomenon known in the literature as
concept drift. We propose a new method for detecting concept drift which uses an
Exponentially Weighted Moving Average (EWMA) chart to monitor the
misclassification rate of an  streaming classifier.
 Our approach is modular
and can hence be run in parallel with any underlying classifier to provide an
additional layer of concept drift detection. Moreover our method
is computationally efficient with
overhead $O(1)$ and works in a fully online manner with no need to store data
points in memory. Unlike many existing approaches to concept drift detection,
our method allows the rate of false positive detections to be controlled and
kept constant over time. 
%\sout{Extensive testing on both artificial and real
%world
%data sets shows that our method provides significant performance
%improvements
%classification tasks with concept drift.}

\end{abstract}
\begin{keyword}
streaming classification, concept drift, change detection


\end{keyword}

\end{frontmatter}

% \linenumbers


\section{Introduction}
\label{sec:introduction}

In many situations it is necessary to analyze data streams consisting of time
ordered data points which are being received at too high a rate, and in too
large volumes, for conventional statistical methods to be deployed. In this case
streaming (online) methods must be developed which are more
computationally efficient. Streaming methods are usually required to meet the
following criteria \cite{Domingos2003}:

\begin{enumerate}
	\item Single pass: points from the data stream should be processed only
once and discarded rather than stored in memory. It may be acceptable to store a
small number of points for repeated processing, but the maximum number stored
should be small and constant rather than increasing indefinitely over time.
	\item Computationally efficient updates: the time required to process
each point should be small and constant over time. The computational complexity
should be $O(1)$.
	%\item Adaptable: streaming methods should be able to cope with changes
%in the behavior of the stream without incurring too severe a decrease in
%performance.
\end{enumerate}

\textbf{Classification} is a common task in the analysis of data streams, where
objects must be assigned to one of several classes based on their observed
features \cite{Hastie2001}. The streaming version of the classification problem has the following form: at
time $t$,
the classifier is presented with the feature vector $\mathbf{f}_t$
of
a single object belonging to an unknown class. It is required to predict the class
of this object. The task is to incrementally learn a classification rule
which assigns objects to classes. In the literature on streaming classification,
it is usual \cite{Baena-Garcia2006, Gama2004, Kuncheva2009} to assume that the
true class of the object is revealed immediately after the prediction is
made so that the classifier receives immediate feedback on whether the
classification was accurate. 
%More formally,
%define a set of $n$ objects which can each belong to one of $k$ classes as:

%$$S = \{(\mathbf{f}_t, c_t)\} \quad t=1,2,...,n$$

%where $\mathbf{f}_t$ is the feature vector which describes the $t$'th
%object and $c_t \in \{l^1,...,l^k\}$ is the class label indicating to which of
%the
%$k$ classes the object belongs. 

The key difference between the streaming
classification problem and the conventional offline version is that, in the
streaming case, the optimal classification rule may change over time due to
changes in the stream dynamics, a phenomena known as \textbf{concept drift}
\cite{Widmer1996}. A distinction can be made between cases where the optimal
rule is gradually changing, and cases where the change is abrupt. For the
remainder of the paper we assume that changes are abrupt, although we will briefly consider gradual drift in our experimental analysis.

In classification tasks where concept drift may occur, it is important to design
classifiers which can adapt to changes in the stream so that they do not
incur a significant decrease in performance. Many methods which have been
proposed to deal with concept drift fall into one of two categories. The first
is to design classifiers which automatically adapt their behavior
to stay
up-to-date with the stream dynamics \cite{Widmer1996,Kolter2007,Kuncheva2008}. Alternatively, classification
and concept drift can be treated as separate problems and concept drift
detectors
are designed to flag when changes occur, and allow some action to be taken \cite{Gama2004,Baena-Garcia2006,Kuncheva2009}. Methods of the second kind are useful in situations where it is not only necessary to adapt to concept drift, but also to give some indication that it has occurred; for example, if classification techniques are used to detect credit card fraud \cite{Viaene2002} it may be necessary to take further investigative action of the behavior of fraudsters is thought to have changed.  In this paper we are concerned only with with the second type of method, and our goal is to detect the change points at which concept drift occurs.
%One popular approach belonging to the first category is to maintain a training
%window containing only the most recent objects \cite{Widmer1996}. The classifier
%is trained using only the objects in this window, ensuring that the classifier
%adapts to the recent behavior of the stream. Choosing the right size for this
%window is difficult when no information is available about how frequently
%changes occur, and several authors have proposed methods of adapting this window
%size online \cite{Klinkenberg2004,Kuncheva2008}. 

%A recent example of an approach in the second category is presented in
%\cite{Gama2004} where the expected misclassification rate is monitored, and it
%is flagged that concept drift has occurred if this rises. However the authors do
%not specify the details of how the misclassification rate is estimated. A more
%complete description is given in \cite{Kuncheva2009}, where a
%working algorithm is presented for concept drift detection. A similar method is
%given in \cite{Baena-Garcia2006} where the time between misclassifications is
%monitored, instead of the misclassification rate.

Most existing approaches to concept drift detection have two main limitations. First, many of them are not single-pass, and have a computational complexity 
which grows with the number of observations. This makes them unsuitable for streaming classification problems where large numbers of 
observations are received frequently. Second, there is generally no way to control the false positive rate, where a false positive is defined as the detector flagging that concept drift has occurred, when in fact there is none. This is a serious problem in cases where it is desirable to know whether concept drift has really occurred. Suppose that a standard concept drift detection method such as \cite{Gama2004,Maloof2008} is used on a stream containing $500$ observations. Suppose also that $5$ different change points are detected, at which abrupt concept drift occurs. Now, we wish to ask whether these are genuine change points, or simply false positives that have been flagged by the detector due to statistical fluctuation. Because with most existing concept drift detectors there is no way of knowing the rate at which false positives are occurring, we do not know whether these change points are likely to be significant: if the detector generates a false positive every $100$ observations, then it is quite likely that there is actually no concept drift, and all the detected instances are false positives. However if we had a way to control the false positive rate so that the detector generates one false positives roughly every (e.g.) $5000$ observations for any data stream, then we could conclude with some degree of certainty that the change points are likely to be genuine instance of concept drift.

%uenerewith large numclassifhigh voluc ttiis points receivdhence may ythey are not 
%The limitation of most of these approaches is that they are not single pass. The
%training window is allowed to grow indefinitely which is not feasible in
%applications where large numbers of data points are being received. Furthermore
%since all previous observations are reclassified at each time point, this
%approach is computationally expensive, with complexity $O(n)$. Therefore it will
%not be feasible when data is received at a high rate. In \cite{Kuncheva2008} an
%alternative approach is presented  based on the Sequential Likelihood Ratio test
%(SLRT) which is single pass and has complexity $O(1)$. The SLRT is used to
%monitor for a change in the misclassification rate with concept drift being
%flagged if it is found to increase. However this approach does not have any
%obvious way of controlling the rate of false positive detections,
%and this rate may fluctuate over time making performance analysis difficult.
%Indeed, the subtle issues of performance assessment of concept drift detectors
%is under-explored at present.
 
In this paper we present an alternative approach to concept drift detection
which is both single pass and computationally efficient with only $O(1)$
overhead, and which allows the rate of false positive detections to be
hence controlled.  We consider the two-class classification problem, although our method could be
extended to the multi-class case. Suppose
we have a streaming classifier which predicts class labels for the objects
$\mathbf{f}_1
\ldots \mathbf{f}_n$. Assuming that feedback is received on whether the
prediction is
correct, we can form the error stream $\{X_t\}$ where $X_t = 0$ if the
prediction for the class of point $\mathbf{f}_t$ is correct and $X_t = 1$ if it
is
incorrect. $\{X_t\}$ can then be viewed as a sequence of observations from a
Bernoulli distribution, with the Bernoulli parameter $p$ corresponding to the
probability of misclassifying a point. Detecting concept drift then becomes the
problem of detecting an increase in $p$, beyond that associated with sampling
variability.

Since our method uses only this
error stream, it treats the underlying classifier as a black-box and does not
make use of any of its intrinsic properties. Therefore, it is able to be
deployed alongside any classifier (decision trees, neural networks, support vector machines, etc)
to provide a modular layer of concept drift detection. This is in contrast to concept drift detectors which are designed
only to work with (eg) linear discriminant classifiers \cite{Kuncheva2008}, or support vector machines \cite{Klinkenberg2000}.

Several schemes for detecting a change in a Bernoulli parameter
have been proposed; however much of this is either not single pass
\cite{Pettitt1980,Bell1994} or assumes the pre-change value of $p$ to be known
\cite{Reynolds1999}. We choose to adapt the Exponentially Weighted Average
(EWMA) chart recently
developed in \cite{Yeh2008} so that it can function in concept drift detection.

This paper proceeds as follows: Section \ref{sec:EWMA} presents some general
background about the EWMA chart,
and Section \ref{sec:bernoulliEWMA} shows how it is applied to the Bernoulli
distribution. The standard formulation of the EWMA chart assumes that various
parameters of the stream being monitored are known. Section \ref{sec:framework}
explains how the EWMA can be adapted for practical situations where these  are
unknown. Section \ref{sec:controllimits} presents a
method for keeping the rate of false positives constant over time. Section
\ref{sec:ECDD} describes how to incorporate this EWMA chart into a concept drift
detector, and we name our algorithm ECDD (EWMA for Concept Drift
Detection). Section
\ref{sec:experiments} analyses the performance of our approach on several real
data sets, and compares it to several other recently proposed methods.

\section{Background}
\label{sec:EWMA}

Exponentially Weighted Moving Average (EWMA) charts were originally proposed in
\cite{Roberts1959} for detecting an increase in the mean of a sequence of random
variables. Suppose we observe the independent random variables $X_1,\ldots, X_n$
which have a common mean $\mu_0$ before the change point and $\mu_1$ after. We
write $\mu_t$ for the mean at time $t$, noting that this quantity only has two
possible values. For now it is assumed that both $\mu_0$ and $\sigma_X$, the standard
deviation of the stream, are known. In Section \ref{sec:framework} we will show how to proceed when this is not the case. Define the EWMA estimator of $\mu_t$ as:

\begin{equation}
\label{eqn:ewma}
Z_0 = \mu_{0}
\end{equation}
$$Z_t = (1-\lambda)Z_{t-1} + \lambda X_t, \quad t>0$$

This EWMA estimator is essentially a way of forming a `recent' estimate of
$\mu_t$, with older data being progressively downweighted. The parameter
$\lambda$ controls how much weight is given to more recent data compared with
older data. It can be shown \cite{Roberts1959} that, independent of the
distribution of the $X_t$
variables, the mean and standard deviation of $Z_t$ are: 

$$\mu_{Z_t} = \mu_t, \quad \sigma_{Z_t} =
\sqrt{\frac{\lambda}{2-\lambda}(1-(1-\lambda)^{2t})}\sigma_X.$$

Before the change point we know that $\mu_t = \mu_0$, and the EWMA estimator
$Z_t$ will fluctuate around this value. When a change occurs, the value of
$\mu_t$ changes to $\mu_1$, and $Z_t$ will react to this by diverging away from
$\mu_0$ and towards $\mu_1$. This can be used for change detection by flagging
that a change has occurred when:

\begin{equation}
\label{eqn:flagchange}
Z_t > \mu_0 + L \sigma_{Z_t}.
\end{equation}

The parameter $L$ is called the \textbf{control limit} and determines how far
$Z_t$ must diverge from $\mu_0$ before a change is flagged. The value of $L$ is
normally chosen to ensure that the detector achieves some predefined level of
performance. A common performance measure is the expected time between false
positive detections, denoted
$\mathbf{ARL_0}$ (for Average Run Length). A false positive here is defined
as the EWMA chart flagging that a change has occurred when $\mu_t$ has not
changed. $L$ is chosen so that the expected
time between false positives is equal to some desired value for $ARL_0$.
Determining which value of
$L$ corresponds to a desired $ARL_0$ is non-trivial, and will be discussed later
in Section \ref{sec:controllimits}.

\subsection{The Bernoulli EWMA for Change Detection}
\label{sec:bernoulliEWMA}

Suppose there is an online classifier which predicts class labels for the observations with feature vectors
$\mathbf{f}_1, \ldots, \mathbf{f}_n$. Assuming that immediate feedback is
received on whether the
prediction is correct, let $\{X_t\}$ be the error stream as defined in Section
\ref{sec:introduction}. This error stream can be viewed as a sequence of
Bernoulli random
variables with the Bernoulli parameter $p_t$ representing the probability of
misclassifying a point at time $t$. Detecting concept drift then reduces to the
problem of detecting an increase in the parameter $p_t$ of a Bernoulli
distribution. 
Again
it is assumed that $p_t$ has only two possible values: $p_0$
before the change point and $p_1$ after, although this is a slight idealization
as will be discussed further in Section \ref{sec:experiments}. We note here in
passing that concept drift may occur without
affecting the error rate, but these situations will be very rare, and  since
classification performance is not affected, detecting the concept drift is not
paramount. Therefore we will not consider these cases further, and assume
throughout that concept drift results in an increased error rate. 

A EWMA change detector for the Bernoulli distribution
was considered in \cite{Yeh2008}, under the assumption that $p_0$
and
$\sigma_{X}$ are known in advance. When working with the Bernoulli distribution,
$\sigma_X$ now depends on $p_t$, so that any change in the $p_t$ will also
change
the standard deviation. To make this explicit we add a subscript and assume that
$\sigma_{X_t} = \sigma_0$ before the change point, and	$\sigma_{X_t} =
\sigma_1$ after.

If the EWMA estimator $Z_t$ is defined as in the previous section, then
elementary properties of the Bernoulli distribution give the
pre-change standard deviation of the EWMA estimator as \cite{Yeh2008}:

\begin{equation}
\label{eqn:ewmavar}
\sigma_{Z_t}^2 =\sqrt{ p_0(1-p_0) \frac{\lambda}{2-\lambda}(1-(1-\lambda)^{2t})}
\end{equation}

\section{Concept Drift Detection}
\label{sec:framework}
The above approach needs several modifications before it can be used for
streaming concept
drift detection. The main problem is that it is assumed that $p_0$ is known,
whereas in practical streaming classification problems
this will not be the case and it must instead be estimated from the stream
along with $\sigma_0$. Therefore, in addition to the above EWMA estimator
$Z_t$, we introduce a second estimator of $p_0$ which we
denote by $\hat{p}_{0,t}$, defined as:

$$\hat{p}_{0,t} = \frac{1}{t}\sum_{i=1}^{t}X_i =  \frac{t-1}{t}\hat{p}_{0,t-1}
+ \frac{1}{t}X_t.$$

 Unlike $Z_t$,
the $\hat{p}_{0,t}$ estimator does not
give more weight to recent observations from the stream. This implies that $Z_t$
is more sensitive to changes in $p_0$, and should give an estimate close to
its \textbf{current} value. The $\hat{p}_{0,t}$ is less sensitive to
changes in $p_0$, and is therefore intended to be an estimate of its
\textbf{pre-change} value.

When a change in the value of $p_0$ occurs, the $Z_t$ estimator should react
more
quickly
and converge towards the new value.  The $\hat{p}_{0,t}$ estimator should 
converge towards this new value more slowly. Our EWMA procedure flags
for a change whenever the distance between these two estimators exceeds a
certain threshold, i.e. when $$Z_t > \hat{p}_{0,t} + L \sigma_{Z_t}$$ where we
have simply substituted the estimate $\hat{p}_{0,t}$ for the known quantity
$p_0$ in Equation \ref{eqn:flagchange}. The pre-change standard deviation can then be
estimated by $$\hat{\sigma}_{0,t} = \hat{p}_{0,t}(1-\hat{p}_{0,t}).$$
Substituting into Equation \ref{eqn:ewmavar} gives the standard deviation of the
EWMA estimator as:

$$\sigma_{Z_t} =
\sqrt{\hat{p}_{0,t}(1-\hat{p}_{0,t})\frac{\lambda}{2-\lambda}(1-(1-\lambda)^{2t}
)}$$

Estimating $p_0$ online also has implications for the choice of the control
limit, $L$. It is desirable for a change detection algorithm to have a constant
rate of false positives --- a false positive should be equally likely to occur
at any point of the stream. In other words, the $ARL_0$ should preferably be constant
through time. However, determining which value of $L$ will give a desired
$ARL_0$ is only possible if the standard deviation $\sigma_{X_t}$ of the stream
is known, which in turn depends on knowledge of $p_0$. When $p_0$ is unknown,
the control limit must instead be chosen based on the estimate $\hat{p}_{0,t}$.
However this estimate will vary over time, which means that in order to keep the
expected rate of false positives constant, the value of the control limit must
be recomputed every time $\hat{p}_{0,t}$ is updated. This implies that $L$ must
now vary over time, so we add the subscript $L_t$. We propose a method of
varying this control limit in Section \ref{sec:controllimits}.

The final EWMA parameter to be chosen is the value of $\lambda$, with the usual
recommendation \cite{Basseville1993} being to choose $\lambda \in [0.1,0.3]$.
The optimal value of $\lambda$ will depend on the pre- and post-change values
of $p_t$. Since these will usually not be known in advance, it is more
important to choose $\lambda$ to give good performance over a wide range of
concept drift detection problems. We have found that a value of $\lambda=0.2$
is suitable for this purpose.  This is investigated further in
Section \ref{sec:experiments}.

\subsection{The Choice of Control Limits}
\label{sec:controllimits}
Having estimated the parameters required to set up a EWMA chart, the final
design stage is to choose the control limit $L$. In the change detection
literature the usual procedure for choosing $L$ is to decide on an acceptable
mean rate of false positive change detections ($ARL_0$), where an $ARL_0$ of $\gamma$ implies that a false positive is generated every $\gamma$ observations on average, and then to choose $L$ to
achieve this rate. Unfortunately, there is no easy procedure for determining
which value of $L$ corresponds to a required $ARL_0$.

The inverse problem of finding the $ARL_0$ corresponding to a given value of
$L$ can be solved by either an approach based on integral equations
\cite{Basseville1993} or by Monte Carlo techniques \cite{Verdier2008}. One
possible method for choosing $L$ to achieve a desired $ARL_0$ is to conduct
a Monte Carlo search where the $ARL_0$ of various choices of $L$ are evaluated
until one is found that gives a $ARL_0$ close enough to the required value of
$L$.
Generally, this procedure is computationally expensive. However in the case
where $p_0$ is known this is not a major problem since the computation only
has to be performed once. Therefore, this search can be carried out before
monitoring of the stream begins and no computational overhead is added to the
change detector.

When $p_0$ is unknown the problem is more complicated. As discussed in Section
\ref{sec:framework}, obtaining a constant rate of false positives is only
possible if we allow the control limit to be time varying and hence add the
subscript $L_t$. In order to use the above method to determine $L_t$, the Monte
Carlo search would need to be carried out at every time instance whenever $p_t$
is updated, which is likely to be too computationally expensive in practice.

In \cite{Sparks2000} a solution to this problem was proposed for a different
change detection method (the Cumulative Sum chart) and we propose to adapt their method for use
with our EWMA detector. Suppose $f(p_0;ARL_0)$ is the function which returns the
value of $L$ corresponding to a desired $ARL_0$ for some value of $p_0$. The
general idea in is to approximate this function by a
polynomial, using standard regression techniques to estimate the coefficients.
Although this approximation is computationally expensive, it again only needs
to be performed a single time, and so it can be carried out before monitoring
of the stream begins. Therefore no overhead is added to the concept
drift detection.

We are essentially generating a `look-up table' which contains the values of
$L_t$ which give a required $ARL_0$ for various values of $p_0$. Then once
stream monitoring begins,  we can simply use this table to find the required
value of $L_t$ for the current estimate $\hat{p}_{0,t}$ which is an $O(1)$
operation and extremely fast.
We generate the polynomial approximations as follows: for a given value of the
$ARL_0$, compute the values of $L$ corresponding to various values of $p_0$ in
the range $[0.01,1]$ using the Monte Carlo approach from \cite{Verdier2008}.
Regression can then be used to fit a degree $m$ polynomial to these values, of
the form $L = c_0 + c_1 p_0 + \ldots + c_m p_0^m$. Our results show that a
degree $7$ polynomial is adequate to give an accurate fit.

The fitted polynomial approximations for several values of the $ARL_0$ are given
in Table \ref{tab:FAPestimates}, when $\lambda=0.2$. Similar tables for other
values of $\lambda$ can be easily derived. As an example of how this
table is used, suppose that
it is desired to maintain a rate of $1$ false positive per $1000$ data points,
so $ARL_0 = 1000$. If at time $t$, $\hat{p}_{0,t} = 0.1$, then the value $0.1$
is substituted into the appropriate polynomial in the table to give the required
value of $L_t$ at time $t$. 

We note that since the functions simply map the estimated value $\hat{p}_{0,t}$ to the required control limit, they can be used for any choice of base classifier and data stream; there is no need to recompute these functions for each particular monitoring task. The type of classifier used is also not important; any classifier
which (e.g.) has a misclassification rate of $p_0 = 0.1$ will have the same
threshold assigned by this table. Different classifiers will produce different error rates and hence have different values for the threshold parameter, but this mapping can be done using Table \ref{tab:FAPestimates}.

\begin{table}
\begin{center}
\small{
\begin{tabular}{|l|l|}
		  \hline
			 $ARL_0$ & Regression Estimate of L \\
			\hline 	
100 & $2.76 - 6.23\hat{p}_0 + 18.12\hat{p}_0^3 - 312.45\hat{p}_0^5 +
1002.18\hat{p}_0^7$ \\
400 & $3.97 - 6.56\hat{p}_0 + 48.73\hat{p}_0^3 - 330.13\hat{p}_0^5 +
848.18\hat{p}_0^7$ \\		 
1000 & $1.17 + 7.56\hat{p}_0 - 21.24\hat{p}_0^3 + 112.12\hat{p}_0^5 -
987.23\hat{p}_0^7$ \\
			\hline 
		\end{tabular}
}
		\end{center}
	\caption{Polynomial approximations for $L$ for various choices of
$ARL_0$ and $\lambda=0.2$}
	\label{tab:FAPestimates}
\end{table}

\section{The Complete ECDD Algorithm}
\label{sec:ECDD}
We now present our complete algorithm for detecting concept drift, which we call
ECDD (EWMA for Concept Drift Detection). Given a streaming classification
problem, first choose both the classifier to use, a desired value for the
$ARL_0$, and a value for $\lambda$ which we will later assume to be $0.2$. We
show evidence in Section \ref{sec:experiments} that the choice of $\lambda$ is
not critical.
Objects in
the stream are sequentially presented to the classifier, and at each time point
define $X_t = 0$ if the predicted class label was correct, and $X_t = 1$ if it
was incorrect. The estimates $\hat{p}_{0,t}$, $\hat{\sigma}_{X_t}$ and
$\hat{\sigma}_{Z_t}$  are updated using $X_t$. Next, a polynomial from Table
\ref{tab:FAPestimates} is used to find the value of the control limit $L_t$ 
which gives the desired $ARL_0$ for the current estimate of $p_0$. The EWMA
estimator $Z_t$ is updated, and if $Z_t > \hat{p}_{t,0} + L_t\sigma_{Z_t}$ then
it is flagged that concept drift has occurred. Action will then usually be taken
to modify the classifier in response to this, but the details of which action to
take depends on the particular classifier being used. For the rest of paper, we
will assume that the classifier is completely reset, with all previous data being
discarded. It must then be relearned using the data after the change point, beginning with the next observation.

Pseudo-code for this algorithm is given in Table \ref{fig:algorithm}.

\begin{table}
\begin{center}
\fbox{
    \begin{minipage}{9 cm}
\begin{tabbing}

Choose a desired value for $\lambda$ and the $ARL_0$ \\

Initialize the classifier \\

$Z_0 = 0$ and $\hat{p}_{0,0}=0$ \\

For \= eac\=h object $f_t$ \\

\>	classify object and update classifier \\
	
\>	Define $X_t = 0$ if the object was correctly classified \\
\> \> or $X_t$ if the classification was incorrect,\\
	
\>	$\hat{p}_{0,t} = \frac{t}{t+1}\hat{p}_{0,t-1} + \frac{1}{t+1}X_t$ \\
	
\>	$\hat{\sigma}_{X_t} = \hat{p}_{0,t} (1-\hat{p}_{0,t})$ \\
	
\>	$\hat{\sigma}_{Z_t} =
\sqrt{\frac{\lambda}{2-\lambda}(1-(1-\lambda)^{2t})}\hat{\sigma}_{X_t}$ \\
	
\>	Compute $L_t$ based on current value of $\hat{p}_{0,t}$ \\
\>\> using  Table \ref{tab:FAPestimates} \\
	
\>	$Z_t = (1-\lambda)Z_{t-1} + \lambda X_t$ \\
	
\>	Flag for concept drift if $Z_t > \hat{p}_{0,t} + L_t\hat{\sigma}_{Z_t}$
\\
\end{tabbing}

	\caption{Final ECDD algorithm}
	\label{fig:algorithm}
	\end{minipage}
}
\end{center}
\end{table}	

\subsection{Warning Threshold}
\label{sec:warningthresholds}
In the above presentation of our algorithm, we suggested that the classifier should be completely reset whenever concept drift is detected. However in practice we can often do better than this; because an abrupt change in the stream will usually take some time to be detected, the most recent observations which came before the point at which we flagged for change will generally come from the post-change rather than the pre-change distribution. If we store a small number of the most recent observations in memory, then we can train the classifier on these after it has been reset, to give it a headstart compared to beginning the training process with the observation following the change point. Although storing points violates the single-pass assumption of our method, in practice only a very small ($ < 10$) number need to be stored in order to give a performance increase, so this will generally not be a problem.

Recall that we flag for concept drift if $Z_t > \hat{p}_{0,t} + L_t\hat{\sigma}_{Z_t}$. We now introduce a second threshold $W_t$ called the \textbf{warning threshold}, which we define as $W_t = 0.5L_t$. Then, if $Z_t > \hat{p}_{0,t} + W_t\hat{\sigma}_{Z_t}$, we treat this as a warning that concept drift may have occurred and that the detector is about to flag for it. After this warning has been given, subsequent observations from the stream are retained in memory. If concept drift is then flagged, these observations are used to retrain the classifier. If instead $Z_t$ later drops below this warning threshold, then we conclude that this warning was false and that no concept drift occurred, and the stored observations are discarded. We will refer to the implementation of our algorithm which uses warning thresholds as ECDD-WT

We note that there is a slight degree of arbitrariness about our choice of $0.5$ for the warning threshold, since other values could also have been used. This specific choice was based on empirical experiments, and we show in the next section that it gives either equal or improved performance compared to the basic ECDD algorithm across all the datasets we consider. However it may be possible that in some situations a different value of the threshold would be reasonable. The value of W implicitly represents a belief about how long any occurring concept drift will take to be detected. Suppose the concept drift occurs at time $\tau$, and ECDD detects it at time $\hat{\tau}$. If $\hat{\tau} - \tau$ is very small, which corresponds to the concept drift being detected very soon after it occurs, then a relatively high value of W should be used since very few of the observations $x_{\hat{\tau}},x_{\hat{\tau}-1},\ldots$ will be from the pre-change distribution. Similarly, if $\hat{\tau} - \tau$ is large then W should be relatively low, since most of recent observations $x_{\hat{\tau}},x_{\hat{\tau}-1},\ldots$ will be from the pre-change distribution. Therefore, since large magnitudes of concept drift will generally be detected quickly, we can say that W should be high if it is suspected that the magnitude of change will be large (which corresponds to gross changes in the class distributions, label switching, etc), and low if the magnitude of change is small, which corresponds more to gradual drift. We picked $0.5$ as a compromise between these two extremes.

\section{Experiments}
\label{sec:experiments}
We now assess the performance of the ECDD and ECDD-WT algorithms on several synthetic and
real world data sets, and compare it to two alternative algorithms for
concept
drift detection which can also be deployed alongside any base classifier: the Paired Classifier
(PC)
method described in \cite{Maloof2008}, and the Sequential Probability
Ratio Test (SPRT) method described in \cite{Kuncheva2009}.

We evaluate performance using two different base classifiers: the streaming
Linear Discriminant Analysis (LDA) classifier described in
\cite{Anagnostopoulos2009}, and K-Nearest Neighbours (KNN),with $k=3$. LDA is
chosen
since it is computationally
inexpensive, and can be written in a recursive form which makes it very suitable
for streaming classification. We chose KNN as a simple
example of a classifier which utilizes more complex decision boundaries. Note
that our implementation of KNN is not
recursive, and the class of the $t^{th}$ observation is predicted using the
previous $t-1$ observations in the usual way. There are
adaptations of KNN which make it more suitable to data streams \cite{Law2005},
but since we are only concerned with comparing the performance of
concept
drift detectors, we will not explore this further. For both ECDD and SPRT, we
discarded all old data and reinitialized the classifiers whenever a change was
detected.  We note that the PC algorithm
has a memory facility which allows a small number of observations to be
retained after the change, with the rest discarded. It is hence comparable to our ECDD-WT approach from Section \ref{sec:warningthresholds}.

Both the PC and SPRT detectors have tunable parameters which must be set by the user.
Given a particular data set, the   approach taken in \cite{Maloof2008} is to evaluate the detector on the data set
using many different sets of parameters, and then choose only the set which give
the
best performance. We feel this is a slightly unrealistic approach to take, and
prefer to find a small set of parameter values which gives
acceptable performance on a wide range of data sets. 

As we have done throughout
the paper, we view the change detection problem in terms of the mean time
between false positives ($ARL_0$). If we
choose a set of parameters which gives false positives every $100$ observations
on average, then it will generally detect concept drift faster than one which
gives false positives every $600$ observations. However the increased number
of false positives will have a negative effect on performance; whenever the
detector flags for a change, most old data from the stream will be discarded
and the classifier reinitialized. This will cause performance to drop until
enough new observations have been seen to allow the classification rule to be relearned.

To investigate this, we decided to use two versions of each concept drift
detector with $ARL_0$'s of $100$ and $600$ respectively, in order to test how
classification performance is affected. These particular values were chosen since (as seen below) they are equal to $2 \times T$, where $T$ is the location of  
the change point in the artificial data sets which we consider. We would expect detectors using an $ARL_0$ of $100$
 to give better performance on streams where changes occur early, and
the detectors with an $ARL_0$ of $600$ to give better performance when changes occur after many
observations. For our ECDD algorithm, we used the
 appropriate polynomial from Table
\ref{tab:FAPestimates} to select the control limit for ECDD with $\lambda=0.2$. We stress again that due to the way this Table has been constructed, 
the same polynomial can be used for each data set and will give the target $ARL_0$.

Tuning the SPRT and PC methods to give a required $ARL_0$ is a more difficult
problem, since these approaches do not contain any way of adapting their
parameters online in order to control the false positive rate. The PC
algorithm has two parameters, $w$ which defines the size of a window, and
$h$ which acts as a threshold. Similarly the SPRT algorithm has two parameters
$\alpha$ and $\beta$, which roughly correspond to the probability of
making a Type I and Type II error.

However there is no obvious way to choose values for these parameters without knowing features of the data stream in advance.
  For a fair comparison with our algorithm, we chose values which gave an $ARL_0$ of $100$ and $600$. However unlike with our approach, 
the parameters which give these values for the false positive rate vary depending on the data set and base classifier.
 For example we found that using the PC detector, the parameter values which give an $ARL_0$ of $600$ on the SINE dataset using the LDA
 classifier only give an $ARL_0$ of $285$ on the GAUSS dataset using the same classifier. Because there is hence no way to control the 
false positive rate in advance, it is difficult to assess the statistical signifiance of any change points found using these approaches, 
as discussed in Section \ref{sec:introduction}.

%\begin{table}
%\begin{center}
%\scriptsize
%\begin{tabular}{|c|c|c|}
%%\hline
%\multicolumn{3}{|c|}{LDA} \\ \hline
%Dataset  & PC & SPRT \\
%\hline
%GAUSS:& a=0.18,b=0.12 (400), a=0.18,b=0.12 (100) & a=0.18,b=0.12 (400), a=0.18,b=0.12 (100) \\
%SINE: &  a=0.18,b=0.12 (400), a=0.18,b=0.12 (100) & a=0.18,b=0.12 (400), a=0.18,b=0.12 (100) \\
%STAGGER &  a=0.18,b=0.12 (400), a=0.18,b=0.12 (100) & a=0.18,b=0.12 (400), a=0.18,b=0.12 (100) \\
%\hline
%\multicolumn{3}{|c|}{KNN} \\ \hline
%Dataset  & PC & SPRT \\ \hline
%GAUSS:& a=0.18,b=0.12 (400), a=0.18,b=0.12 (100) & a=0.18,b=0.12 (400), a=0.18,b=0.12 (100) \\
%SINE: &  a=0.18,b=0.12 (400), a=0.18,b=0.12 (100) & a=0.18,b=0.12 (400), a=0.18,b=0.12 (100) \\
%STAGGER &  a=0.18,b=0.12 (400), a=0.18,b=0.12 (100) & a=0.18,b=0.12 (400), a=0.18,b=0.12 (100) \\
%\hline
%\end{tabular}
%%\caption{Parameter values for the PC and SPRT classifiers which give the desired false positive rate, in brackets.}
%\label{tab:paramvalues}
%\end{center}
%\end{table}

Finally, we note that both our ECDD, ECDD-WT and the SPRT algorithms have a very low
computational overhead, with only a small number of calculations being
performed at each time step. The computational overhead of the PC algorithm is
much higher, unless the underlying classification algorithm can be written in a
special form 
\cite{Maloof2008}, which limits the situations in which it can be deployed.

\subsection{Artificial Data Sets}
We evaluate performance on two artificial data sets
containing abrupt changes which are widely used as benchmarks in the concept
drift literature \cite{Gama2004,Kuncheva2008}. Both of these contain two classes:

\textbf{GAUSS}: Before the change, points with class label 0 are drawn from a
bivariate
Gaussian distribution with mean vector $(0,0)^T$ and identity covariance $I$,
and
points
with class label 1 are drawn from a Gaussian distribution with mean vector
$(2,0)$ and
covariance matrix $4I$. After the change point these classifications are
reversed.

\textbf{SINE}: Data set with two independent features. Both features are
uniformly distributed on [0,1]. Before the change point, all points below the
curve $y=\sin(x)$ are class 0, and points above are class 1. This classification
reverses after the change point.

The time of the change point will affect performance, since a change which
occurs early in the stream will be harder to detect as the relevant parameters
(the error rate $p_0$ in the case of our EWMA algorithm) will not yet be
accurately estimated. To take this account, we use two versions of the GAUSS
and SINE data sets, with the change points occurring at $T=50$ and $T=200$ respectively, We write
GAUSS50 to denote the GAUSS data set with the change occurring after the
$50^{th}$ point, and so on. The length of each stream is $2T$, so there are $400$
total observations in the streams which have a change after $200$ observations, and
$100$ in the streams which have a change after $50$ observations.

%new material.

 For each data set and value of $T$, we generated $10000$ realizations of each
data set and calculated
the average classification accuracy using each base classifier and concept
drift detector. 
%As well as reporting the average accuracy over the entire
%stream, we also report the average classification accuracy on the $20$
%observations immediately following the change point. This gives a clearer
%picture of how well the classifiers are performing around the point where
%concept drift occurs, since the classification accuracy over the whole stream
%will be dominated by the observations before and long after the change point,
%which are comparatively easy to classify.

We first investigate the effect that varying the $\lambda$ parameter has on the
ECDD method. In the EWMA literature, it is usual to set $\lambda$ to a value
less than $0.3$, since using a higher value results in too much emphasis being
placed on recent data, making parameter estimation difficult due to the high
variance. We therefore consider the values $\lambda = \{0.1,0.2,0.3\}$.

The results when the change detectors have an $ARL_0=600$ are shown in Table
\ref{tab:lambdasyntheticARL400}, with the results for $ARL_0=100$ being
similar,
but omitted for space reasons. From this, it seems that the ECDD algorithm is
not particularly sensitive to the value of $\lambda$ chosen. Therefore, for the
rest of this section we use the value $\lambda=0.2$.

\begin{table}
\begin{center}
\scriptsize
\begin{tabular}{|l|l|l|l|l|}
\hline
$\lambda$  & Gauss50  & Gauss200  & Sine50 & Sine200 \\
\hline
0.1 &0.59 (0.07)  & 0.73 (0.03) & 0.78 (0.06)& 0.91 (0.02)\\
0.2  &0.60 (0.07)  & 0.72 (0.03) & 0.78 (0.06)& 0.90 (0.02)\\
0.3 &0.60 (0.07)  & 0.72 (0.03) & 0.78 (0.06)& 0.89 (0.02)\\
\hline
\end{tabular}
\caption{Classification accuracy on synthetic data sets using the ECDD
algorithm with several choices of $\lambda$, and $ARL_0=600$, standard errors
shown in brackets.}
\label{tab:lambdasyntheticARL400}
\end{center}
\end{table}

Next, we compare our approach to the PC and SPRT algorithms.  $10000$
realizations of each data set were generated. The results when the change
detectors have $ARL_0 = 100$ and
$ARL_0=600$ are shown in Tables \ref{tab:syntheticARL400} and
\ref{tab:syntheticARL120} respectively.

From these tables, we see that using any concept drift detector gives a large
improvement in performance compared to simply running the base classifiers
without assistance. More interestingly, these results also show the impact of
false
positives on performance. When the change occurs after $200$ observations, the
detectors with an $ARL_0$ of $600$ out-perform those with an $ARL_0$ of $100$.
This
is because although the lower $ARL_0$ allows changes to be detected faster, the
increase in false positives outweighs this benefit. However when the change
occurs after $50$ observations, there is less time for false
positives to occur before the change, and the detectors with a lower $ARL_0$
perform better. This highlights the importance of matching the false positive
rate of the detector to the rate at which changes are occurring in the stream.

Finally, the results show that the ECDD algorithm
gives similar performance to both the PC and SPRT methods. When the warning thresholds are incorporated 
as described in Section \ref{sec:warningthresholds}, the performance of the ECDD approach improves further. We note again that we have not attempted to
optimize the value of the $ARL_0$ for any of the three concept drift detection algorithms, and it may be possible to improve the performance of all methods by considered different
values. However since this kind of optimization will not be possible in practice, we do not pursue it further.

Note that the differences in performance between the various concept drift detectors are generally quite small; 
this is because the detector generally only affects how quickly the change is detected, which will affect classification performance
 only on the few observations following the change point. In order to verify that the differences in classification accuracies were
 statistically significant, we followed standard practice \cite{Dietterich1998} in using Mcnemar's test to make pairwise comparisons
 between the compared algorithms. Due to the high number of simulations used, p-values of less than $10^{-5}$ were obtained in all cases, showing significant results.

\begin{table}
\begin{center}
\scriptsize
\begin{tabular}{|l|l|l|l|l|}
\hline
Classifier  & Gauss50  & Gauss200  & Sine50 & Sine200 \\
\hline
LDA &  0.51 (0.06)& 0.52 (0.03) & 0.50 (0.07) & 0.52 (0.04)  \\
LDA-ECDD & 0.59 (0.07)&0.71 (0.03) & 0.77 (0.06)& 0.90 (0.02)\\
LDA-ECDD-WT &0.60 (0.07)  & 0.72 (0.03) & 0.78 (0.06)& 0.90 (0.02)\\
LDA-PC   & 0.62 (0.05)& 0.71 (0.03)&0.79 (0.06)& 0.90 (0.02)\\
LDA-SPRT   & 0.58 (0.04)& 0.72 (0.04) & 0.78 (0.05)& 0.91 (0.02)\\ 
\hline
KNN  & 0.54 (0.06)& 0.57 (0.03)& 0.54 (0.06)& 0.62 (0.03)\\
KNN-ECDD & 0.61 (0.07)&0.73 (0.03) &0.79 (0.08) & 0.91 (0.01)\\
KNN-ECDD-WT & 0.62 (0.07)&0.73 (0.03) &0.79 (0.07) & 0.92 (0.01)\\
KNN-PC   &  0.63 (0.05)& 0.73 (0.03)& 0.78 (0.06)& 0.91 (0.01)\\
KNN-SPRT   & 0.61 (0.04)& 0.73 (0.04) & 0.80 (0.05)& 0.92 (0.01)\\
\hline
\end{tabular}
\caption{Classification accuracy on synthetic data sets for concept drift
detectors with $ARL_0=600$, standard errors shown in brackets.}
\label{tab:syntheticARL400}
\end{center}
\end{table}

\begin{table}
\begin{center}
\scriptsize
\begin{tabular}{|l|l|l|l|l|}
\hline
Classifier  & Gauss50  & Gauss200  & Sine50 & Sine200\\
\hline
LDA &  0.51 (0.06)& 0.52 (0.03) & 0.50 (0.07) & 0.52 (0.04)  \\
LDA-ECDD &0.63 (0.07) & 0.71 (0.03)& 0.79 (0.06)&0.89 (0.03)\\
LDA-ECDD-WT & 0.64 (0.07) & 0.70 (0.03)& 0.80 (0.06)&0.89 (0.03))\\
LDA-PC   & 0.64 (0.05)& 0.70 (0.03)& 0.78 (0.06)& 0.89 (0.02)\\
LDA-SPRT   & 0.63 (0.04)& 0.70 (0.04)& 0.80 (0.05)& 0.89 (0.02)\\
\hline
KNN  & 0.54 (0.06)& 0.57 (0.03)& 0.54 (0.06)& 0.6 (0.03)\\
KNN-ECDD &0.65 (0.07) & 0.72 (0.03)& 0.80 (0.07)&0.90 (0.02)\\
KNN-ECDD-WT & 0.66 (0.07) & 0.72 (0.03)& 0.81 (0.07)&0.90 (0.02)\\
KNN-PC   &  0.65 (0.05)& 0.72 (0.03)& 0.78 (0.06)& 0.89 (0.01)\\
KNN-SPRT   & 0.65 (0.04) & 0.71 (0.04) & 0.81 (0.05)& 0.90 (0.01)\\
\hline
\end{tabular}
\caption{Classification accuracy on synthetic data sets for concept drift
detectors with $ARL_0=100$, standard errors shown in brackets.}
\label{tab:syntheticARL120}
\end{center}
\end{table}

\subsection{Gradual Drift}
In the experiments in the previous section we assumed that the concept drift consisted of abrupt changes. However in some situations, it may be the case that concept drift is caused by gradual change. Although our algorithm, like the SPRT and PC methods we have been comparing it to, is primarily intended to be used to detect abrupt change, it is important to investigate whether it can still give acceptable performance when gradual drift is encountered.

Unfortunately, the majority of standard synthetic concept drift benchmark data sets contain only abrupt changes rather than gradual drift. The exceptions are the rotating hyperplane dataset, and one which consists of moving circles \cite{Gama2004}. However in these datasets, the concepts are in a continual state of drift and there is no period when they are static. In this situation none of the algorithms we have considered would be expected to perform well, and a classification ensemble would be more suitable \cite{Kolter2007}.

We therefore instead choose to modify the GAUSS and SINE datasets to produce a stream which is initially static, and then undergoes gradual drift for a period of time.  In the experiments in the previous section, the true class label of each observation was immediately switched following the change point. We now make the modification that after the change point, each observation has probability $q_t$ of having its label switched. The previous case was equivalent to $q_t=0$ before the change point, and $q_t=1$ after. We now allow $q$ to gradually drift from $0$ to $1$. We assume that the change point occurs at time $T=200$, and that $q_t$ increases linearly from $0$ to $1$ over the interval $[200,300]$ to simulate moderate drift. The classification performance using an $ARL_0$ of $400$ is shown in Table \ref{tab:drift}.

\begin{table}
\begin{center}
\scriptsize
\begin{tabular}{|l|c|c|}
\hline
Classifier  & Drifting GAUSS  &Drifting SINE\\
\hline
LDA-ECDD & 0.68 & 0.86\\
LDA-ECDD-WT & 0.69 & 0.86 \\
LDA-PC  &  0.69 & 0.87\\
LDA-SPRT  & 0.68 & 0.85\\
\hline
KNN-ECDD&0.69 & 0.82\\
KNN-ECDD-WT&0.69 & 0.82\\
KNN-PC     & 0.70 & 0.87\\
KNN-SPRT   & 0.69 & 0.80\\
\hline
\end{tabular}
\caption{Classification accuracy on the drifting synthetic data sets. Standard errors shown in brackets.}
\label{tab:drift}
\end{center}
\end{table}

It can be seen that our ECDD algorithm again gives competitive performance for this type of gradual drift, being equal to the SPRT in most cases. However
both methods are outperformed by the PC approach, suggesting that it is the better choice when concept drift may take the form of gradual drift instead of abrupt change.
However this performance advantage must be balanced against the presviously mentioned limitations of this method, namely its high computational overhead, and the inability to control the rate of false positives.

% We note however that the for all three methods, improvement over the base classifier performance is not as large as it was when changes were abrupt. This is because when the drift is detected, the base classifier is reinitialised from scratch with all previous data immediately forgotten. Although this procedure is ideal when abrupt change occurs, it may not be the best approach for gradual drift since if they drift continues for some time, there may be multiple such resettings of the classifier. One alternative would be to use gradual forgetting \cite{Kuncheva2009} rather than complete forgetting but we do not pursue this further.

\subsection{Real Data}
Finally, we analyze two real world data sets:  the Electricity Market data which is
standard in the concept drift literature \cite{Harries1999}, and a set of data
related to colonoscopic imaging. With both data sets, we do not know in advance whether
concept drift is present, or what form it takes if it exists (abrupt changes
versus gradual drift). As before, the data sets are
treated as if they were streams and classification is performed in an
incremental sequential manner.

With real data sets, choosing the parameter values used in the SPRT and PC concept drift detectors is a serious problem. With the previous
artificial data sets we had knowledge about the location of the true change point, and could use this to determine the parameter values which gave a desired $ARL_0$.
However with the Electricity data we do not know where the true change points are, so this is not possible. There is therefore no obvious way of controlling the 
$ARL_0$ of the PC and SPRT detectors, and this is the key limitation of these methods and the primary advantage of ours.

As a compromise, we have evaluated these concept drift detectors on the data sets using a wide variety of parameter settings, and have reported the performance when using 
the set which gives optimal performance. Therefore, the results below give an indication of the best possible performance for each method. Because this is slightly unrealistic, and in practice it will not generally be possible to finely tune these methods in such a manner, we also report the performance which is achieved when the parameters are varied over a range of values. For our ECDD detector, this involves letting the $ARL_0$ range from $100$ to $1000$. For the PC and SPRT detectors this is more difficult, since there are several free parameters. We therefore report their performance over a range of values centered around the empirical best value.

\subsubsection{Electricity Data Set}
The data used for this comparison is a set of prices collected from the New
South Wales Electricity Market as described in \cite{Harries1999}. The prices
from this market were logged at 30 minute intervals between 7 May 1996 and 5
December 1998, giving a total of 45312 feature vectors. Each feature vector
contains 5 features: the time at which the sample was taken, the NSW electricity
demand,  the Vic electricity demand, the scheduled electricity transfer between
states, and the class label. The class label is $0$ if the price has increased
compared to a moving average taken over the last 24 hours, and $1$ if it has
stayed the same or decreased.

We tackle the problem of predicting the price movement over each 30 minute
period using only the NSW and Vic demands available on that day, which gives a
two class classification task with two features. This is a simple model, which
ignores possible autocorrelation and seasonal trends in the data, but it is
sufficient for our purposes.

The data set is classified both with and without
concept drift detection, and the overall classification accuracy is shown in
Table \ref{tab:elec}. It can be seen that incorporating concept drift detection
gives a significant performance increase for both LDA and KNN approaches.
Greater accuracy is obtained using KNN, suggesting that the optimal
classification boundary is non-linear. Interestingly, the performance of all three methods is identical, assuming the best parameter settings are used for each. 

In order to investigate the effect of using nonoptimal parameter settings, we allowed the $ARL_0$ of the ECDD method to vary from $100$ to $1000$. The optimal classification accuracies of $0.86$ and $0.88$ for the LDA/KNN classifiers respectively were obtained when the $ARL_0$ was $100$. When the $ARL_0$ is increased to $1000$, these accuracies gradually drop to $0.85$ and $0.87$. This implies that performance is quite robust, with the $ARL_0$ not being overly critical (within reason). Interestingly, the fact that the best performance is achieved with a low $ARL_0$ suggests that changes are occurring quite frequently. Results for the SPRT and PC classifiers were very similar; when the parameters were varied in a small range around their best values, performance dropped only by a very small amount. 

\begin{figure}[ht]
\centering
\includegraphics[scale=0.4]{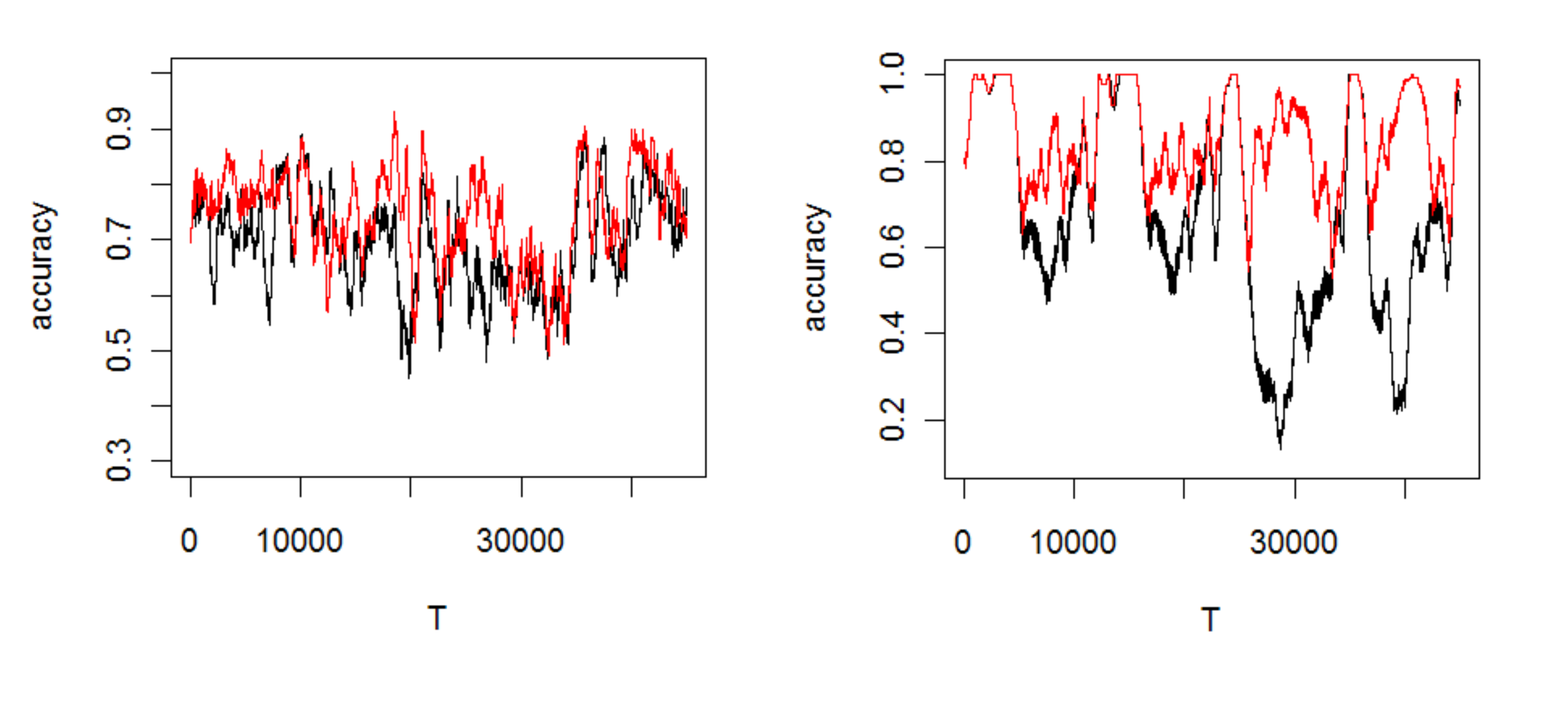}

\caption{Classification accuracy of the Streaming LDA classifier on the
Electricity (left) and Colonoscopic (right) data sets over time, using ECDD
(red) and performing no change
detection (black)}
\label{fig:elecerror}
\end{figure}

Figure \ref{fig:elecerror} shows how the average classification accuracy changes
over time for the LDA classifier when using ECDD compared with not performing
any concept drift detection. This was computed by moving a sliding window of
size $100$ over the data, and using the average accuracy over the points
$[t,t+99]$ as an estimate of the error rate at time $t$. From this graph it
appears that the accuracy when using ECDD is higher over most of the data set.
Further investigation would be required to determine whether this is because the
data set contains abrupt change points which we are detecting, or whether we are
detecting the accumulation of gradual drift.

\begin{table}
\begin{center}
\scriptsize
\begin{tabular}{|l|l|l|}
\hline
Classifier  & Electricity & Colon \\
\hline
LDA &  0.70 & 0.68 \\
LDA-ECDD &  0.86 & 0.90 \\
LDA-PC  &  0.86 & 0.90 \\
LDA-SPRT  &  0.86 & 0.90 \\
\hline
KNN &  0.73 & 0.89 \\
KNN-ECDD &  0.88 & 0.90\\
KNN-PC  &  0.88 & 0.90 \\
KNN-SPRT  &  0.88 & 0.90 \\
\hline
\end{tabular}
\end{center}
\caption{Classification accuracy of the streaming LDA and KNN
classifiers on
the Electricity and Colon data sets for various concept drift detection methods}
	\label{tab:elec}	
\end{table}

\subsubsection{Colonoscopic Video Sequencing}

The accurate online classification of imaging
data from colonoscopic video sequences can contribute to the early
detection of colorectal cancer precursors, and
assist in the early diagnosis of colorectal cancer. We obtained a sample of one
of these imaging data sets. In this dataset textures
from normal and abnormal tissue samples were randomly chosen from four
frames
of the same video sequence without applying any preprocessing to the
data~\cite{KarkanisMT2000}. Feature extraction was performed using the
method
of co-occurrence matrices~\cite{Haralick1979}. This method represents the
spatial distribution and the dependence of the grey levels within a
local area
using an image window of size 16 by 16 pixels. The final data set contains 17076
feature vectors, each with 16 features. The class label designates whether a
window contains tumor pixels (class 1), or not (class 0). 
The overall classification accuracy for this data set is given in Table
\ref{tab:elec}, and again it can be seen that all methods appear to give the same performance when using their optimal parameter settings. As before, we also varied the parameter settings of the three detectors in order to test their robustness. For our ECDD method, we again found that best performance was achieved when the $ARL_0$ was set to $100$, corresponding to accuracy rates of $0.88$ and $0.90$ for the LDA and KNN classifiers respectively. These gradually dropped to $0.87$ and $0.89$ as the $ARL_0$ was increased to $1000$, suggesting robustness. Similar results were found for the PC and SPRT methods. Again, the fact that the best performance is achieved for a small $ARL_0$ value suggests that changes are occurring frequently.

Figure \ref{fig:elecerror} shows how the average classification accuracy
changes over time using the ECDD algorithm with a LDA classifier. From
this it appears that the colon data set is broken up into several segments of
reduced performance, with an accuracy of close to $1$ between these segments.
This suggests that the data does contain abrupt changes, although further
analysis would be required to verify this.

%Figure \ref{fig:colonerror} shows how the average classification accuracy
%changes over time for the LDA classifier. From this it appears that the colon
%data set is broken up into several segments of reduced performance, with an
%accuracy of close to $1$ between these segments. This suggests that the data
%does contain abrupt changes, although further analysis would be required to
%verify this. 

\section{Conclusions}
We presented ECDD, a method for detecting concept drift in streaming
classification problems based on the Exponentially Weighted Moving Average
chart. Since our method uses only the classification error stream, it can be
incorporated into any streaming classifier assuming that feedback is received
regarding whether predictions are correct. Our approach does not require any
data to be stored in memory, and only adds $O(1)$ overhead to the classifier.
Additionally, it allows the rate of false positive concept drift detections to
be controlled in a manner which other approaches do not. Experimental analysis showed that the performance of our approach 
is competitive with other state of the art methods. One
possible direction of future research is to extend our methodology to
classification problems which have more than two classes, perhaps by monitoring
each entry of the confusion matrix separately.

%\section{Acknowledgements}

%This research was undertaken as part of the ALADDIN (Autonomous Learning Agents
%for Decentralised Data and Information Systems) project and is jointly funded
%by a BAE Systems and EPSRC (Engineering and Physical Research Council)
%strategic partnership, under EPSRC grant EP/C548051/1. David Hand was
%partially supported by a Royal Society Wolfson Research Merit Award.

%% References


\bibliographystyle{elsarticle-harv}
\bibliography{jabref}






% \end{thebibliography}

\end{document}